# A SLAM Map Restoration Algorithm Based on Submaps and an Undirected Connected Graph

Zongqian Zhan[1], Wenjie Jian[1,*], Yihui Li[1], Xin Wang[2] and Yang Yue[1]

[1]School of Geodesy and Geomatics, Wuhan University, China; zqzhan@sgg.whu.edu.cn, wjjian@whu.edu.cn, yhli@whu.edu.cn, 2016301610337@whu.edu.cn
[2]Leibniz University Hannover Institute of Geodesy; wang@ipi.uni-hannover.de
*Correspondence : wjjian@whu.edu.cn ; Tel. : +86-18361227556

**Abstract**: Many visual simultaneous localization and mapping (SLAM) systems have been shown to be accurate and robust, and have real-time performance capabilities on both indoor and ground datasets. However, these methods can be problematic when dealing with aerial frames captured by a camera mounted on an unmanned aerial vehicle (UAV) because the flight height of the UAV can be difficult to control and is easily affected by the environment. For example, the UAV may be shaken or experience a rapid drop in height due to sudden strong wind, which may in turn lead to lost tracking. What is more, when photographing a large area, the UAV flight path is usually planned in advance and the UAV does not generally return to the previously covered areas, so if the tracking fails during the flight, many areas of the map will be missing. To cope with the case of lost tracking, many visual SLAM systems employ a relocalization strategy. This involves the tracking thread continuing the online working by inspecting the connections between the subsequent new frames and the generated map before the tracking was lost. The new frames are then localized under the coordinate system of the generated map if these corresponding connections are found. However, there is a fatal drawback in the relocalization strategy, in that the state of lost tracking will remain until the camera moves back to the place before the UAV lost tracking and the tracking and mapping is restarted. As a result, a part of the map will be missing, from the place where the tracking was lost to the place where the camera can be relocalized. To solve the missing map problem, which is an issue in many applications (e.g., 3D position information for emergency relief), after the tracking is lost, based on monocular visual SLAM, we present a method of reconstructing a complete global map of UAV datasets by sequentially merging the submaps via the corresponding undirected connected graph. Specifically, submaps are repeatedly generated, from the initialization process to the place where the tracking is lost, and a corresponding undirected connected graph is built by considering these submaps as nodes and the common map points within two submaps as edges. The common map points are then determined by the bag-of-words (BoW) method, and the submaps are merged if they are found to be connected with the online map in the undirect connected graph. To demonstrate the performance of the proposed method, we first investigated the performance on a UAV dataset, and the experimental results showed that, in the case of several tracking failures, the integrity of the mapping was significantly better than that of the current mainstream SLAM method. We also tested the proposed method on both ground and indoor datasets, where it again showed a superior performance.

**Keywords:** monocular visual SLAM; UAV images; undirected connected graph; submaps; complete global map

## 1. Introduction

Simultaneous localization and mapping (SLAM) is a technique for obtaining 3D geometric information of an unknown environment and estimating the sensor pose in the corresponding environment. As such, the SLAM technique has a very wide application potential in automatic driving, augmented reality, virtual reality, mobile robots, and unmanned air vehicle (UAV) navigation [1]. With the ongoing development of sensor and computer vision technology, many kinds of sensors have been integrated in SLAM systems (such as LiDAR, GPS, and inertial measurement unit (IMU) sensors) [2]. However, only SLAM based on cameras has been actively studied because the sensor configuration is much simpler. SLAM based on stereo cameras has been widely used with both indoor and ground datasets [3]; however, this setup is not suitable for UAVs (especially micro aerial vehicles) since the length of the baseline is normally only 20 cm, which is quite short when the images are captured from a height of nearly 1 km [4]. SLAM based on a single camera, which is also called monocular visual SLAM, has received extensive attention and has been widely studied for its simplicity and cheapness. Thus, we developed the proposed method using only monocular visual SLAM. In monocular visual SLAM, without the assistance of any other sensors, lost tracking often occurs for many reasons, such as image blurring caused by the camera moving too fast (UAV platforms, in particular, are quite easily affected by random shaking caused by uncertain factors such as wind), illumination variation, and weak scene texture features. Once the tracking of a monocular visual SLAM system (such as ORB-SLAM2) is lost, the motion of the subsequent video frames and the information of the corresponding map points cannot be generated, which leads to an incomplete map. Figure 1 depicts a set of UAV data, where there were three interruptions during the flight. Figure 1a is UAV data from Wuhan University. Figure 1b shows the flight trajectory of the UAV and the locations of the interruptions (marked by the red rectangles). Figure 1c is the processing result of the ORB-SLAM2 system, where it can be seen that the processing result of ORB-SLAM2 is only partial, due to the interruptions.

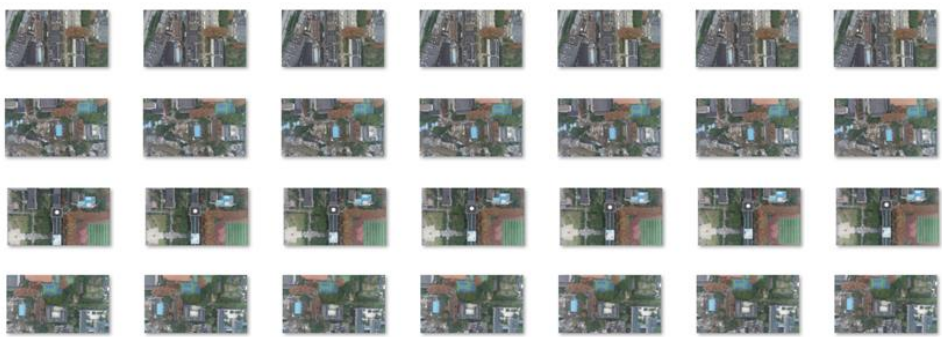

(**a**)

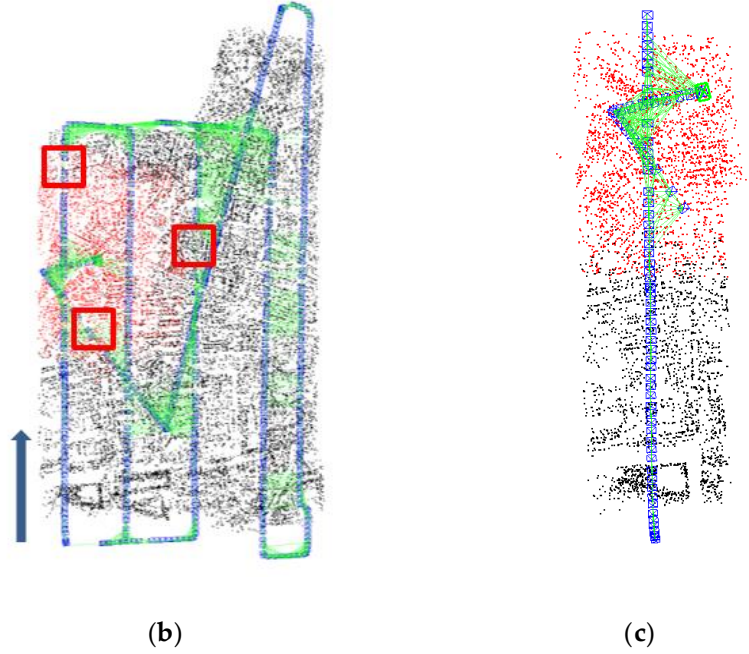

(**b**)            (**c**)

**Figure 1.** UAV dataset experimental results. (a) The UAV data from Wuhan University. (b) The flight trajectory of the UAV and the locations of the interruptions (the red rectangles mark the interruptions). (c) The processing result of ORB-SLAM2.

To solve the lost tracking problem, the current visual SLAM systems employ a relocalization strategy, which makes the tracking thread continue the online working by inspecting the connections between the subsequent new frames after tracking was lost and the generated map before tracking was lost, and localizes the new frames under the coordinate system of the generated map if the corresponding connections are found. However, there is a crucial disadvantage to this relocalization strategy, i.e., the map information and the corresponding camera motion knowledge are not estimated and will be missing from the place where the tracking failed to the place where the new frame could be successfully relocalized. To recover this missing information, including the corresponding map and the camera motion when the unexpected tracking was lost in monocular visual SLAM and provide a complete global motion trajectory and map, which is often required in many applications (e.g., 3D position information for emergency relief), we present a method of recovering the missing map parts using submaps and an undirected connected graph. This is integrated with the local mapping of the tracking or the loop closing thread based on the widely known ORB-SLAM2 package [5]. A general overview of the proposed system is shown in Figure 2. Our main contributions are threefold. Firstly, when lost tracking occurs, we generate a corresponding submap and directly begin new map initialization and continue the tracking and mapping. Secondly, we present a method to reconstruct a complete global map and camera motion trajectory by using the generated submaps and the corresponding undirect connected graph. Finally, we demonstrate the proposed method's performance via evaluation on UAV images, and some ground and indoor datasets are further tested to further show the method's capability.

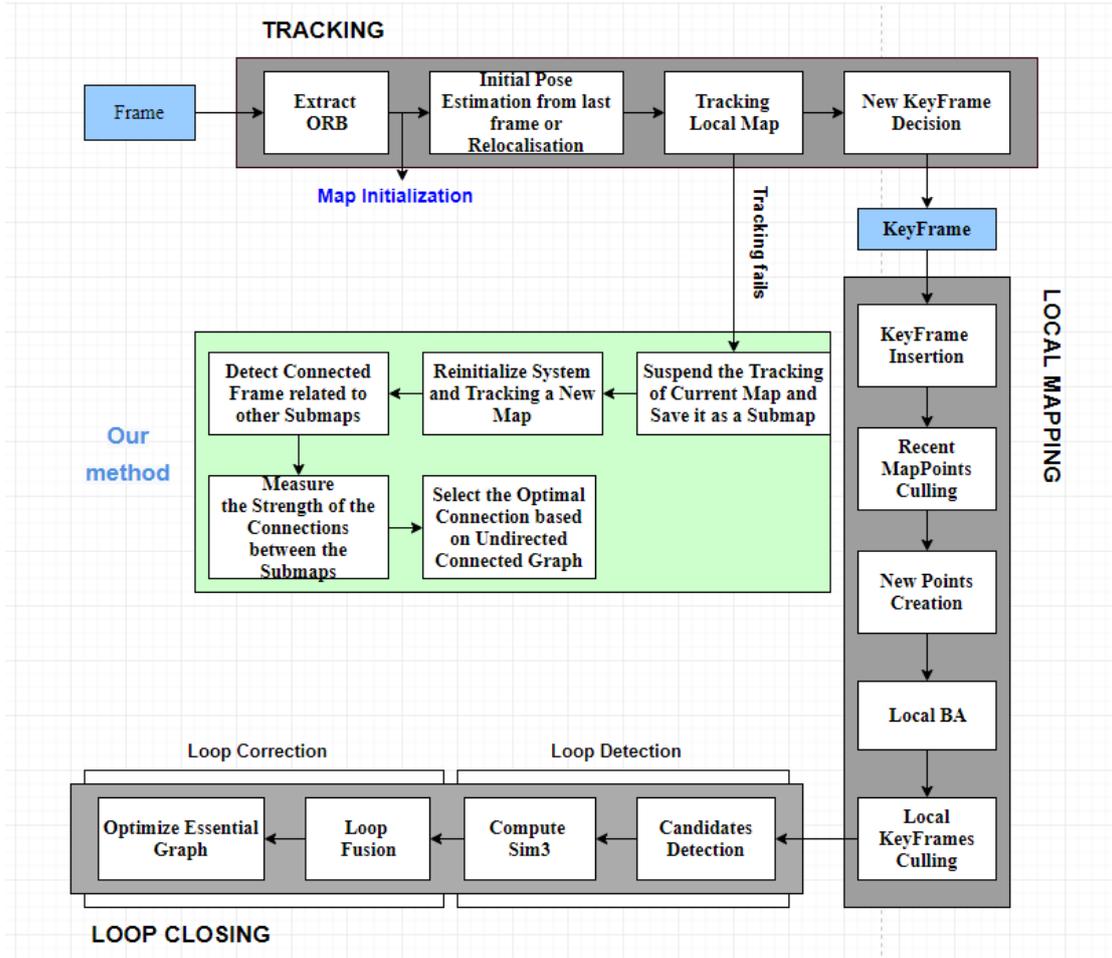

**Figure 2.** System threads and module.

The rest of this paper is organized as follows. In Section 2, we review the related works. Section 3 describes the proposed method of reconstructing a complete global map and camera motion trajectory. Section 4 presents the results of the experiments conducted on various datasets. Finally, Section 5 concludes the paper.

**2. Related Works**

Over recent decades, real-time visual SLAM has been broadly investigated in various fields, such as automation and robotics, computer vision, and photogrammetry. In this section, we review some of the state-of-the-art works in visual SLAM research. Specifically, we focus on the studies of motion tracking and the solutions for lost tracking and submaps in SLAM

*2.1. Solutions for Motion Tracking*

There has been a great deal of research on improving tracking performance and attempting to decrease the possibility of lost tracking. In visual SLAM, these works can be generally classified into three categories: feature-based methods, direct methods based on photometric consistency, and multi-sensor aided tracking methods.

2.1.1. Feature-based methods

Oriented brief (ORB) features are oriented multi-scale features from accelerated segment test (FAST) corners with 256-bit descriptors, which are widely used in feature-based SLAM, mostly because they are extremely fast to compute and have good invariance to the viewpoint [6]. However, the corresponding processes of feature point extraction and matching can easily be negatively affected by the change of illumination, view angle, and weak texture, which leads to lost tracking. In order to relieve this drawback and improve the tracking performance, other types of features, such as plane features and line features, have been integrated with visual SLAM.

For example, Lee et al. [7] used only plane features in the tracking, which is an approach that is mostly applicable to an environment dominated by plane features; Taguchi et al. [8] presented a method combining point and plane feature primitives to obtain a minimal set of primitives in a RANSAC framework to robustly compute the correspondences and estimate the sensor pose; Raposo et al. [9] adopted plane features as the primitives in visual odometry, and point features were only extracted when the plane features were insufficient to determine the sensor motion; and Concha et al. [10] proposed an approach based on superpixel region matching, which was shown to be more reliable in tracking than the traditional point-feature-based methods for indoor scenes with weak texture and rare features, but this approach does suffer from a limited accuracy in complex outdoor scenes.

Some studies [11-13] have used different line segment parameterization approaches and have tried to use the line features for solving the motion tracking, while essentially employing the two endpoints to describe and track a line segment. Jeong et al. [14] used both 3D line and corner features as landmarks in tracking under an extended Kalman filter framework. Based on ORB-SLAM2, Pumarola et al. [15] presented a line segment detector method to achieve the extraction and matching of straight lines with the same name, combined with ORB features for the tracking, which turned out to be more reliable in scenes with abundant line features and rare point features. Lee et al. [16] extracted the missing points with lines to improve the tracking accuracy and robustness in an indoor scene.

2.1.2. Direct methods based on photometric consistency

The extraction and matching of feature points can be time-consuming, and a scene with weak texture and only very rare features may be detected, which may give a low accuracy or even failure in pose estimation. What is more, these feature points only represent a very small part of the image information, which can be improved by considering all the information given in the image. Thus, the direct methods based on global pixel information assume that the image intensity of the same spatial point should be consistent in the corresponding neighboring images, and the position and orientation of the camera are estimated through minimization of the photometric error. Direct sparse odometry (DSO) [17] combines a fully direct probabilistic model (by minimizing a photometric error) with consistent, joint optimization of all the model parameters, including the geometry, represented as the inverse depth in a reference frame, and the camera motion. In contrast to the feature-based methods, the direct methods can utilize all the information in the image and have a higher tracking accuracy and robustness in scenes with only very rare features. Dense tracking and mapping (DTAM) [18] involves selecting frames and then computing detailed textured depth maps to generate a surface patchwork and build dense maps by GPU acceleration, which are then used for the tracking by comparing the new frames with these dense maps. This approach effectively reduces the uncertainty of pose estimation, and many semi-dense algorithms based on edge and corner features have been proposed. For example, semi-direct visual odometry (SVO) [19] involves extracting the FAST feature points in the image and then estimating the camera pose transformation by the direct method, according to the information around the feature points. Instead of feature extraction, the large-scale direct SLAM method (LSD-SLAM) [20] computes the depth of the semi-dense points with abrupt gradient changes, such as edges and corners, on the basis of SVO, and it uses the same idea as DTAM for the tracking, which is improved by considering the geometric consistency and loop closure. As a result, LSD-SLAM can deal with scenes with a weaker texture and larger scale, and can be run on a CPU in real time.

2.1.3. Multi-sensor aided tracking methods

To improve the tracking performance, some methods have attempted to integrate multiple sensors into visual SLAM. For example, Leutenegger et al. [21] presented a vision and IMU data combination algorithm, in which the camera pose is computed and optimized by marginalization, which has contributed to the rapid development of multi-sensor fusion algorithms; the VINS-Mono method, which was developed by Qin et al. [22], involves embedding a low-cost IMU into visual-inertial odometry, where a tightly coupled, nonlinear optimization-based method is presented to fuse the IMU and feature observations, which can obtain absolute pose estimation and reduce the risk of lost tracking; and Bu et al. [23] proposed a real-time mosaicking system for UAV video by fusing GPS data, so that the camera's pose in the WGS84 coordinate framework can be estimated without ground control points. However, GPS signals are easily obscured on the ground, so the improvement of robustness for SLAM tracking on the ground is still limited.

*2.2. Solutions After Tracking is Lost*

The above tracking methods normally work well; however, tracking can often be lost in practice, e.g., through rapid motion change, poor textures for the feature-based methods, illumination changes for the direct methods, and GPS-denied environments for methods using GPS or IMU signals. Currently, to deal with lost tracking and ensure that the tracking thread works for a long period of time, most of the monocular visual SLAM methods start relocalization detection after tracking is lost to determine the current pose of the camera [4,17,19,22,24]. Relocalization strategies, which are similar to detection of loop closure, can be roughly divided into two types: image-to-image and image-to-map.

The image-to-image methods use a visual BoW model to describe the images by combining word bags and feature points, with the basic rationale being that correspondences should be with the same word bag. Thereby, the similarity between images can be efficiently determined, and after matching similar images, the corresponding relative positions can be solved by using either a five-point algorithm for an essential matrix or an eight-point algorithm for a fundamental matrix. The relocation problem can thus be considered as an image retrieval problem (a similar problem exists in structure-from-motion (SfM) [25-27]) by employing random forest and a hash image retrieval method. The image-to-image relocalization strategy is employed in filtering-based SLAM [28] and the state-of-the-art frame-based SLAM [24, 29]. After similar images are found, the pose of the camera can be recovered only if there are enough correspondences between the current image and the previous ones, so that the system is able to continue tracking and mapping under the coordinate system of the previously reconstructed map.

The idea of image-to-map matching is to determine the connection between the reconstructed map and the current new frame. Specifically, they are connected if there are enough reconstructed map points that can be observed by the current new frame, and vice versa. In the filtering-based SLAM framework, Williams et al. [30] proposed a three-point-pose algorithm combined with the RANSAC algorithm to determine the position and pose of the current camera relative to the map. In the frame-based SLAM framework, Straub et al. [31] proposed to match the descriptors of the current frame with the descriptors associated with the map points stored in the map, to estimate the pose of the current frame. However, image-to-map matching involves a large amount of calculations and is slow. Therefore, Moteki et al. [32] proposed the method of selecting an image-to-image or image-to-map method based on the geometric model between the current frame and the target frame, which is a more efficient approach.

The above-mentioned relocalization strategies have been used in some widely known visual slam systems, such as ORB-SLAM2, which have obvious defects when dealing with the case of tracking failure, and require the current new frame to have a high similarity to the reconstructed map or the oriented frames. If a similar scene is not detected after the system tracking is lost, the system will remain in the lost state and cannot continue mapping and positioning until the relocation is successful. As a result, the corresponding map information (consisting of the map from the place tracking was lost to place the relocation was successfully solved) cannot be recovered.

*2.3. Submaps*

The concept of the "submap" for SLAM was developed by Ni et al. [33], and is defined as a local map with a local coordinate system and frames with known relative pose and 3D map points. A global map consists of several overlapping submaps covering different parts of the entire scene.

The submap was originally proposed to efficiently solve the problem of the high computational cost of global optimization with limited computational resources [34,35]. The global map is divided into several overlapping submaps. The submaps are first optimized individually, and then a single submap is taken as a whole for the global optimization. This strategy of using submaps can effectively reduce the computational cost while obtaining near-optimal results [36-38].

The submaps are generated to solve the problem of missing map information, where the map from the place of the previous initialization to the place where the lost tracking occurred forms a new submap. Specifically, when lost tracking occurs, the subsequent new frames are directly reinitialized, and the new frame is tracked in the new coordinate system. The map obtained before lost tracking occurs again is the corresponding new submap. In this way, submaps are recursively generated. The multiple submaps are then connected to form a global map describing the complete scene.

To solve the problem of missing map information due to tracking failure and to provide a complete global map and motion trajectory, we present a map restoration fusion method based on the generated submaps and the corresponding undirected connected graph. The proposed method is based on the monocular ORB-SLAM2 framework. After the system tracking failure, even if the relocation requirements cannot be met, the tracking and mapping thread can still continue, but the corresponding submaps are saved, and are eventually joined together into a complete map. In theory, the proposed method is applicable to all the previous methods (such as SVO, DSO, and ORB-SLAM), and can act as an effective supplement to the existing methods, to ensure the completeness and accuracy of the mapping.

## 3. Methodology

*3.1. Method and Thought*

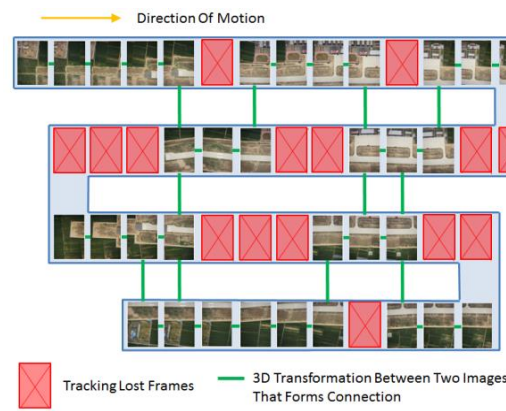

**Figure 3.** UAV trajectory and matching relationship between images.

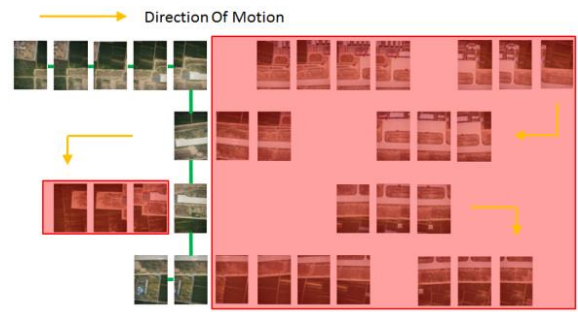

**Figure 4.** The processing result of ORB-SLAM2 (the frames covered by the light red represent the missing maps due to tracking failure).

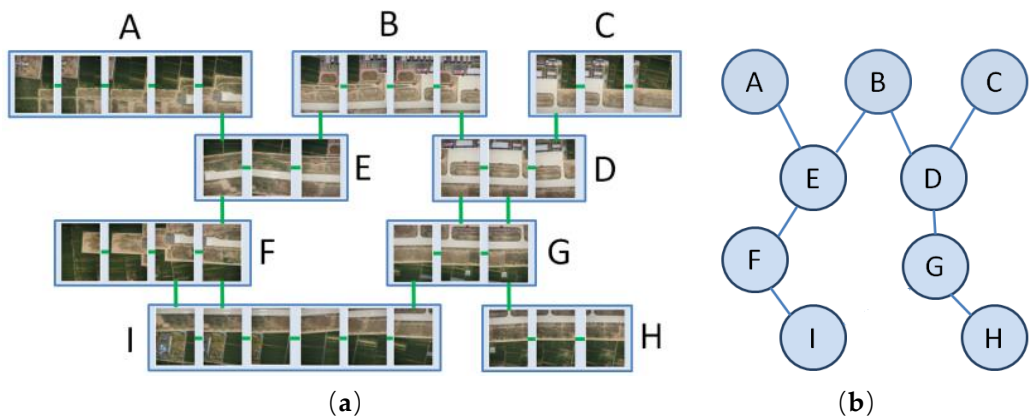

(**a**)            (**b**)

**Figure 5.** (a) The submaps of each segment. (b) The undirected connected graph formed after each submap is connected.

In this section, we introduce the details of the proposed algorithm. Because the motion trajectory of aerial photogrammetry is relatively simple and regular, we used a dataset of UAV video data to illustrate the use of the method proposed in this paper, but the method is also applicable to ground data.

The light blue band in Figure 3 represents the UAV's motion trajectory. The image in the figure represents the frame. The red rectangle indicates that the matching failed at this point and the tracking was lost. The green line indicates that the images can be successfully matched and that there is a connected relationship. Figure 4 shows the processing results obtained by ORB-SLAM2. The frames covered by the light red represent the missing maps due to tracking failure.

In this paper, the red rectangles in Figure 3 are used as intervals, and the frames are grouped according to the acquisition order to obtain Figure 5a. Each set of frames and their corresponding map points constitute a local submap in the global scene map, and are represented as (A, B, C, D, E, F, G, H, I). The system is based on ORB-SLAM2, with the following processes added:

- The system is reinitialized when the tracking fails, and a new map is built.
- Retrieval of connected frames. For the current frame in the map, the BoW model is used to retrieve the frames in the other submaps. Frames are found that match the current frame (which meet a certain threshold), and the retrieved frames are called connected frames.
- The strength of the connections between the submaps is measured.
- The optimal connection based on the undirected connected graph is selected.

*3.2. When the Tracking Fails, the System is Reinitialized to Build a New Map*

The main task of the visual SLAM tracking module is to output the camera poses and determine the frames in real time to complete an unoptimized visual odometer. When the tracking fails, ORB-SLAM2 performs the relocation operation. If the relocation is not successful, all the frames that have not been successfully relocated will be lost until the relocation is successful. In order to avoid the situation of missing maps, when the tracking fails, the previous map is used as a submap, and the system is reinitialized immediately to build a new map.

*3.3. Retrieval of the Connected Frames*

For a newly created map, the system establishes whether the current frame forms a connected relationship with the other submaps. The specific operation is as follows. The adaptive threshold selection method is adopted to retrieve the connected frames. Firstly, the words in the BoW model of the current frame are calculated. The number of words in common between the current frame and the adjacent frame is denoted as $N_0$ (the adjacent frame refers to the previous frame of the current frame), the number of words in common between the current frame and the nearby frame whose overlap with the current frame is about 50% is denoted as $N_1$ (the nearby frame refers to the frame closest to the current frame in the current map), and the proportionality coefficient $k = N_1/N_0$ is calculated. Then, according to the number of common words of the BoW model, all the frames in the other submaps are searched to find the frames with the same words as the current frame. Then, the maximum number of common words of these frames $k$ is set as the threshold, i.e., the frames with an overlap of more than 50% are selected as candidate frames, as shown in Figure 6. The BoW score between the current frame and the adjacent frame is denoted as $S_0$. The BoW score between the current frame and the nearby frames whose overlap with the current frame is about 50% is denoted as $S_1$, and the proportionality coefficient $l = S_1/S_0$ is calculated. Then, the highest score of the BoW score of the current frame and the candidate frames $l$ is then set as the threshold, and the connected frames are then selected based on the BoW score. The function that retrieves the loop frames in the ORB-SLAM2 loop closing thread is then called. The original fixed thresholds (0.8 and 0.75) are then replaced with the previously calculated thresholds, and the frames that meet the conditions as connected frames are finally selected.

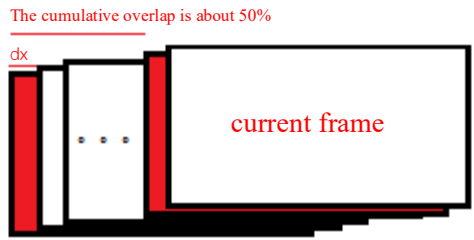

**Figure 6.** The practical significance of the adaptive threshold coefficient. From left to right, the two red frames are the frames that overlap with the current frame by about 50% and the nearest frame, corresponding to the relationship between the actual overlapping range between the nearby frame and the adjacent frame in the threshold selection. The relationship is thus considered, instead of simply setting a fixed absolute parameter.

*3.4. Measuring the Strength of the Connections Between the Submaps*

After a connected frame is found, the map where the current frame is located can form a connected relationship with the submap where the connected frame is located. However, when connecting submaps, there may be multiple connected frames between the submaps. By comparing the geometric configuration of the connected frames, the connection relationship with the highest accuracy and the best reliability for coordinate transformation is selected, so as to combine the multiple submaps.

The geometric configuration of the connected frames mainly refers to the intersection angle between the connected frames. In aerial photogrammetry, the greater the intersection angle of a stereo image pair, the higher the accuracy. When the intersection angle is too small, the error in the direction of the vertical image plane will be very large. In many computer vision SfM 3D reconstruction open-source systems, such as COLMAP, the intersection angle must not be less than 16° when initializing the selected stereo image pair. According to the principle of photogrammetry [39], the smaller the intersection angle of the stereo image pair, the worse the accuracy in the depth direction of the 3D point coordinates obtained from the forward intersection of the stereo image pair, as shown in Figure 7.

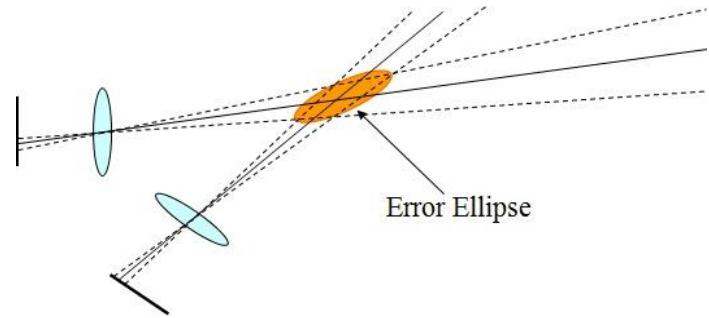

**Figure 7.** Intersection angle and error ellipse.

Since the formulas for photogrammetry are more intuitive than the mathematical models in computer vision, we take the formula for photogrammetry forward intersection as an example. The error of the depth direction is shown in Equation (1):

$$m_h = m_x/\tan\theta ,\qquad(1)$$

where $\theta$ represents the intersection angle of a stereo image pair, $m_x$ represents the plane error, and $m_h$ represents the median error in the depth direction.

$$\tan\theta = \frac{b}{f} = \frac{B}{H} ,\qquad(2)$$

where b represents the image baseline, f represents the camera focal length, b represents the photography baseline, and H represents the photography height, as shown in Figure 8. It is known that if the photographic baseline is too short, the error ellipse will become extremely flat and the depth direction error will be large.

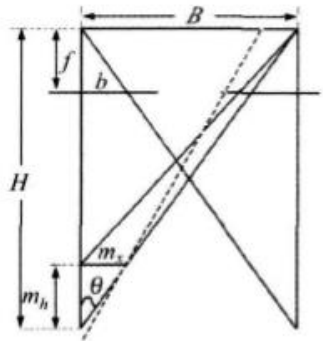

**Figure 8.** The relationship between the intersection angle and depth direction accuracy.

Therefore, if the intersection angle between the connected frames is larger, it indicates that the connection reliability of this part is better, the 3D coordinates of the calculated map points are more accurate, and the error of transforming the two submaps to the same coordinate system is smaller. Of course, the intersection angle should not be too large, and no more than 45° is appropriate. In order to better reflect the number of smaller intersection angles, the geometric configuration of the connected frames is described by the median of the intersection angle, rather than the average.

The number of existing connected frames between two submaps is denoted as F, and the number of map points contained in the connected frame is M. The median intersection angle between connected frames is $\theta$, in degrees. In this paper, a factor C is proposed to measure the connection strength between submaps, as shown in Equation (3). In this study, the empirical values were set through experiments. The order of magnitude F is usually about 10, the order of magnitude M is usually about 100, and θ is usually about 10°. The effect of θ on the submap accuracy is not linear, but instead curvilinear, so this is set to the second power. The strength of the connected frames between the submaps is compared, i.e., the size of the value of C:

$$C = F + 0.1M + 0.1\theta^2 \ . \tag{3}$$

*3.5. Selection of the Optimal Connection Based on the Undirected Connected Graph*

When there are multiple connection paths between multiple submaps, the strongest connection path needs to be selected. This is essentially a problem of finding a connected path for an undirected connected graph. For the connection between submaps, each submap is similar to a group of undirected connected graphs with weights equal to the strength of the different connection paths, as shown in Figure 9. Each submap is a node in an undirected connected graph. The connection relationship between each submap is the edge, and the connection strength corresponds to the weight of each edge. As shown in Figure 5a, the UAV's flight path is A-B-C-D-E-F-G-H-I, and Figure 5b shows the undirected connected graph formed after each submap is connected.

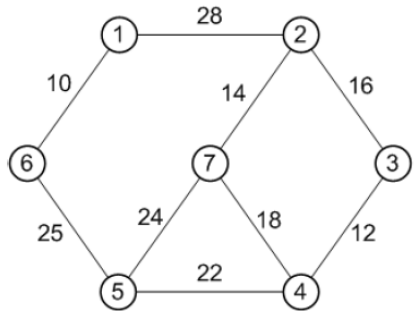

**Figure 9.** An undirected connected graph with weights.

For an undirected connected graph containing n nodes, the process of solving the minimum spanning tree is to find a path that contains n−1 edges and can connect n nodes, so as to minimize the sum of the edge weights (i.e., the cost). For the connection of the submaps, the higher the value of C in Equation (3), the higher the connection strength. Therefore, a negative value of C was chosen as the weight of the undirected connected graph in this study. Kruskal's algorithm can be used to solve the minimum spanning tree.

*3.6. Algorithm Flow*

Each time the system tracking fails, the construction of the map is restarted. The previously established map is called a submap. The proposed method uses the idea of graph theory to connect these submaps into a complete map. In graph theory, each segment of the submap is taken as a node, and the connected relationship between each segment of the submap is taken as the edge. The connectivity detection is used to determine whether there is a connected relationship between submaps. The map currently being tracked is denoted as C, and the n submaps S1 ... Sn generated by the previous tracking failure are stored in stack L. The algorithm flow is as follows:
1. Track the current map C and denote the current frame as F.
2. Determine whether the current map C is successfully initialized. If it has not been completed, continue to initialize C; if it has been completed, go to the next step.
3. Determine whether the orientation of F in C is successful. If the orientation is successful, proceed to the next step. If the orientation fails, perform the following operations: determine whether the system tracking failure conditions are met (several consecutive frames fail to be oriented), and if the system tracking failure conditions are not met, the currently read F is discarded and the next frame is read and recorded as F. Restart step 3. If the tracking failure condition is met, suspend the tracking of C, denote C as $S_{n+1}$ and save it in L, and create a new map to start tracking. Denote it as C and go back to step 1.
4. Determine whether there is an unrecovered submap $S_i$ ($i \in [1, n]$) in the missing submap stack L. If $S_i$ exists, go to the next step; if $S_i$ does not exist, all the submaps in L are retrieved or no submap exists in L. Go back to step 1 and read the new frame.
5. Determine whether $S_i$ and C are connected through F. If $S_i$ and C are connected, proceed to the next step; if $S_i$ and C are not connected, go back to step 1 and read the new frame.
6. Determine whether the connected strength of the two submaps reaches the threshold. If the threshold is met, convert $S_i$ to the C coordinate system and merge the two submaps. Continue to track C, and go back to step 1 and remove $S_i$ from L; if the threshold is not met, continue tracking C, and go back to step 1.

The algorithm flowchart is shown in Figure 10:

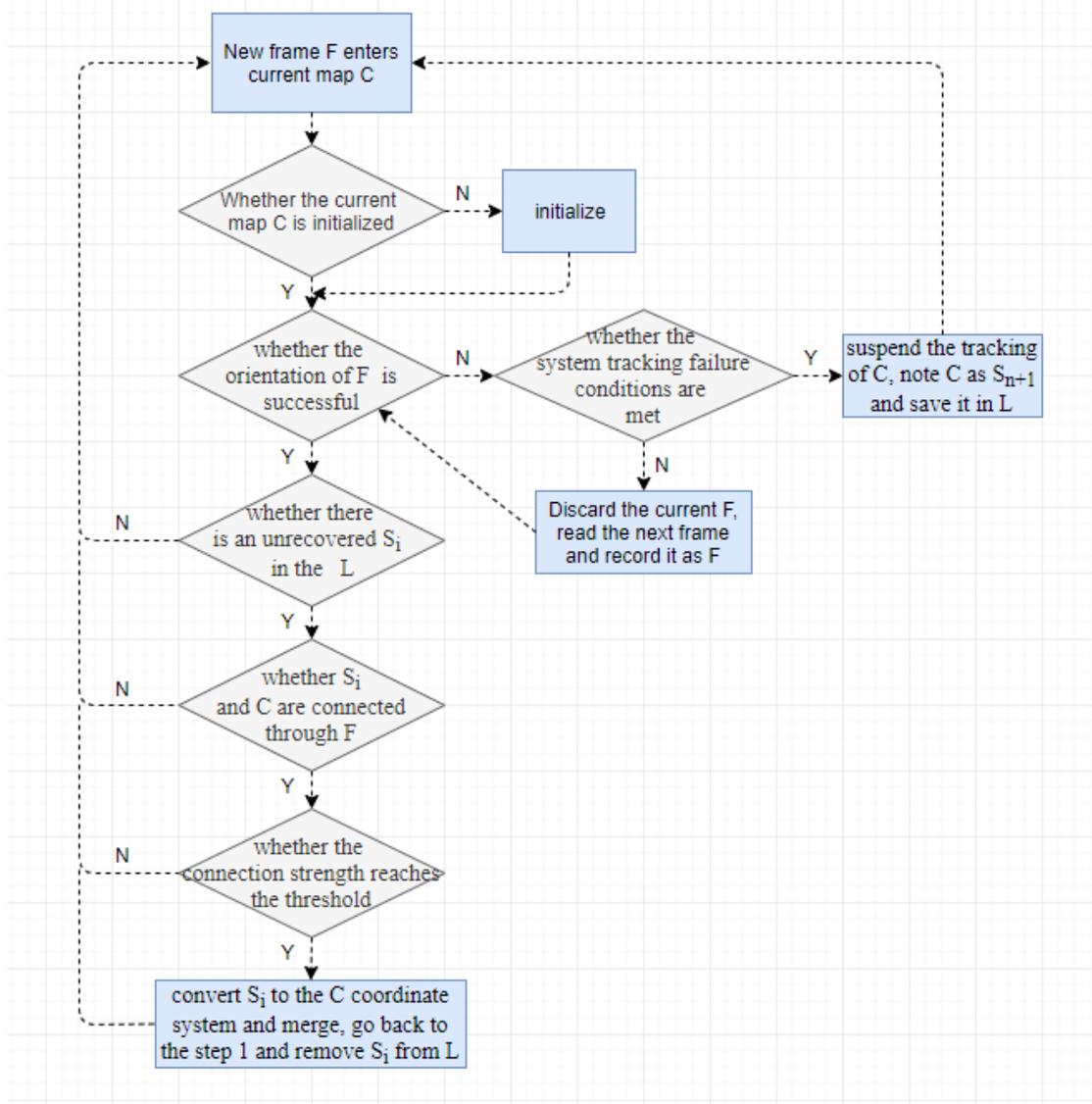

**Figure 10.** Submap algorithm flowchart.

We take Figure 5 as an example to illustrate the process of merging submaps. The new submap is added to stack $L_1$, and the currently tracked map to stack $L_2$. As shown in Table 1, map A is first tracked, and A joins $L_2$; when A fails to track, the system is reinitialized. A, as the first submap, is added to $L_1$, and we start tracking the second map B. $L_2$ is updated to B; after B fails to track, B joins $L_1$, and we start tracking the third map C. $L_2$ is updated to C; after C fails to track, C joins $L_1$, and we start tracking D. $L_2$ is updated to D; when tracking D, the system indicates that D is connected to C, and then merges them into D-C. C is removed from $L_1$, and $L_2$ is updated to D-C. After a while, it is found that D-C is connected to B. These are then merged into D-C-B. B is removed from $L_1$, and $L_2$ is updated to D-C-B, and so on. The submaps are finally merged into I-H-G-F-E-D-C-B-A, which is a complete map. The red letters in Table 1 indicate that the two submaps currently have a connected relationship.

**Table 1.** Submap merging process table.

| Number | $L_1$ | $L_2$ |
|---|---|---|
| 1 | | A |
| 2 | A | B |
| 3 | A, B | C |
| 4 | A, B, C | D |
| 5 | A,B | D-C |
| 6 | A | D-C-B |
| 7 | A, D-C-B | E |
| 8 | | E-D-C-B-A |
| 9 | E-D-C-B-A | F |
| 10 | | F-E-D-C-B-A |
| 11 | F-E-D-C-B-A | G |
| 12 | | G-F-E-D-C-B-A |
| 13 | G-F-E-D-C-B-A | H |
| 14 | | H-G-F-E-D-C-B-A |
| 15 | H-G-F-E-D-C-B-A | I |
| 16 | | I-H-G-F-E-D-C-B-A |

## 4. Experimental Results and Analysis

In order to verify the method proposed in this paper, a UAV dataset, a ground dataset, and two indoor datasets were used in the experiments.

*4.1. UAV Dataset Experiment*

The first dataset is video shot by UAV, as shown in Figure 11. The video resolution is 1920 × 1080, with a total of 16,380 frames. The camera model is the onboard camera of a DJI Phantom 4 drone, and the frame rate is 30 frames per second. The camera orientation is relatively constant, and the angle of view does not change dramatically, being mainly vertically downward. The motion track of the camera is S-shaped. The scenes shot by this group of data are mainly residential areas and vegetation, with abundant feature points.

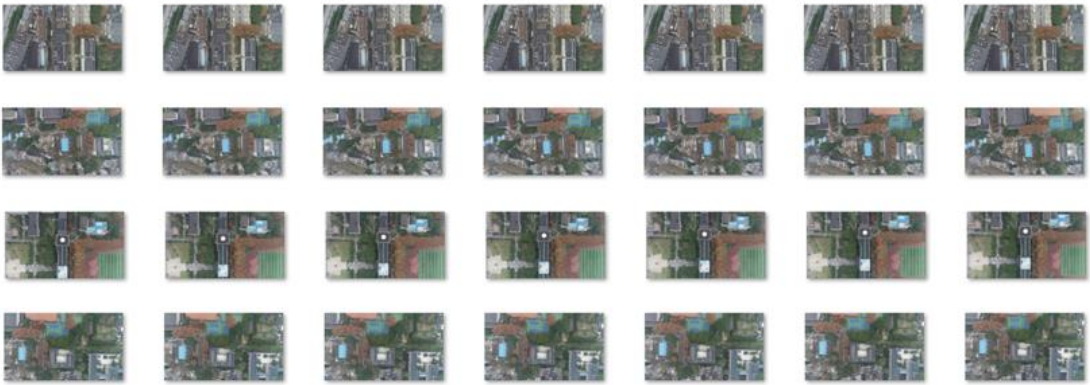

**Figure 11.** UAV experimental dataset.

There were three interruptions during the flight. Figure 12a is the flight trajectory of the UAV. Figure 12b is the processing result of ORB-SLAM2. Figure 12c is the processing result of the proposed system (with the letters representing the submaps). Figure 12d is a schematic diagram of the undirected connected graph. There are three interruptions in this group of data, and the data are divided into four submaps—A, B, C, and D—separated by the interruptions, according to the collection order. When ORB-SLAM2 processed this set of data, it was unable to solve the position of the current frame in the world coordinate system, due to tracking failure between submaps A and B, so submaps B and C were lost, until submap D was successfully relocated and connected to submap A. The final map only includes submaps A and D. After the tracking failure between the A-B and B-C submaps in the proposed system, the new world coordinate system was reinitialized and tracking was continued. At the same time, it was established whether there was a connected relationship between the submaps. Finally, the paths between the submaps A, B, C, and D were detected and restored, and the four submaps were connected in series with the D-A-C-B connected relationship.

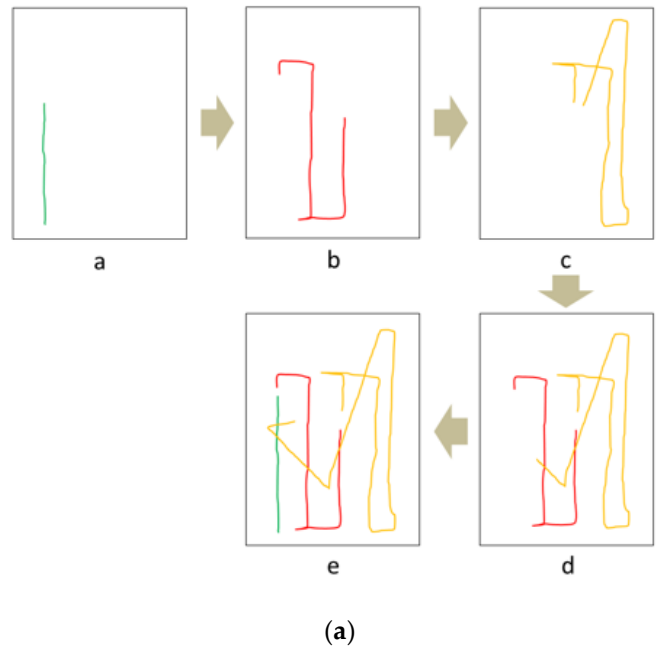

(**a**)

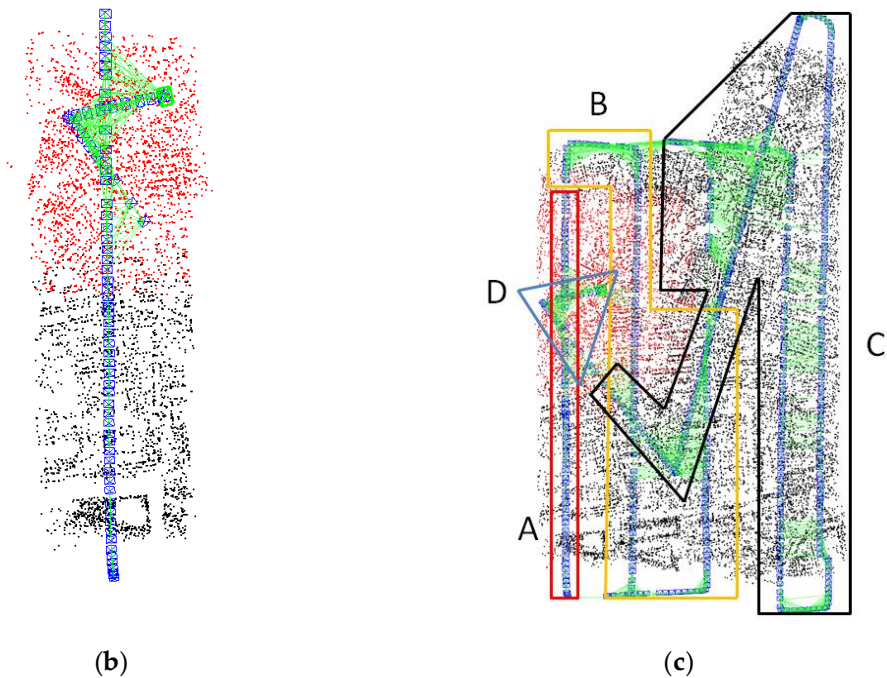

(**b**)        (**c**)

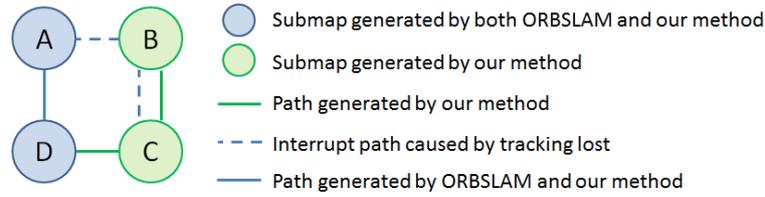

(**d**)

**Figure 12.** UAV dataset experimental results. (a) The flight trajectory of the UAV. (b) The processing result of ORB-SLAM2. (c) The processing result of the proposed system (with the letters representing the submaps). (d) The undirected connected graph.

*4.2. Outdoor Street Dataset Experiment*

The second group of data is videos taken by handheld cameras, as shown in Figure 13. The video resolution is 1920 × 1080, with a total of 24690 frames. The camera model is a GoPro Hero 6 motion camera, with a frame rate of 30 frames per second. The camera orientation is complex and the angle of view changes dramatically. The scenes captured by this dataset are mainly residential streets. The scenes are complex, including moving objects, and areas where it is difficult to extract ORB features. In the process of data collection, the tracking failure was caused by the rapid camera rotation occurring at the street corner.

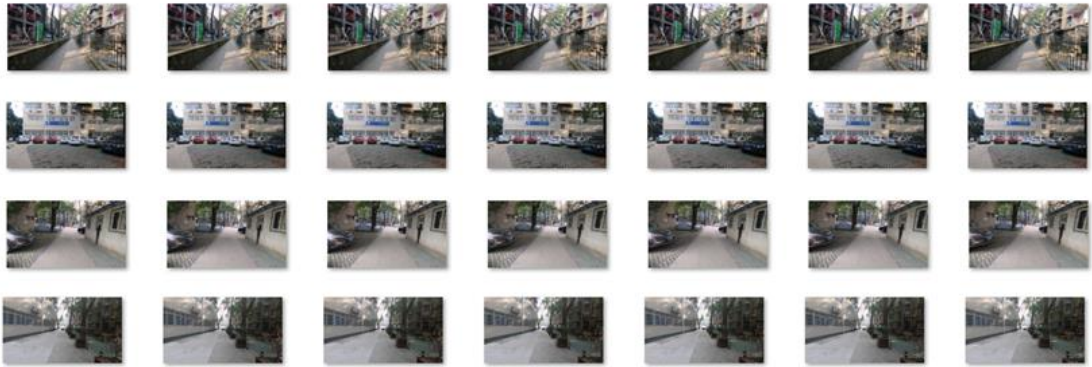

**Figure 13.** Outdoor street dataset.

There are four interruptions at the corners of the street. Figure 14a is the ground running trajectory. Figure 14b is the processing result of ORB-SLAM2. The construction of the map was stopped at the breaks, resulting in many missing maps. Figure 14c is the processing result of the proposed system (with the letters representing the submaps). Figure 14d is a schematic diagram of the undirected connection graph. This is divided into five submaps: A, B, C, D, and E. Among them, A-B and C-D were interrupted due to the tracking failure. C-A and E-C were connected through relocation after the interruption. ORB-SLAM2 processed this data, and the B and D submaps were lost after the tracking failure. However, the proposed method detected the existing connections between B-C and D-E, and restored B and D to the world coordinate system of A-C-E, finally connecting the five submaps.

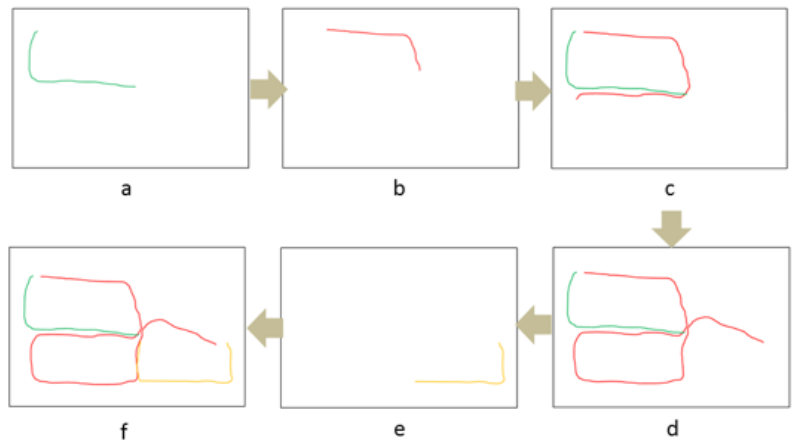

(**a**)

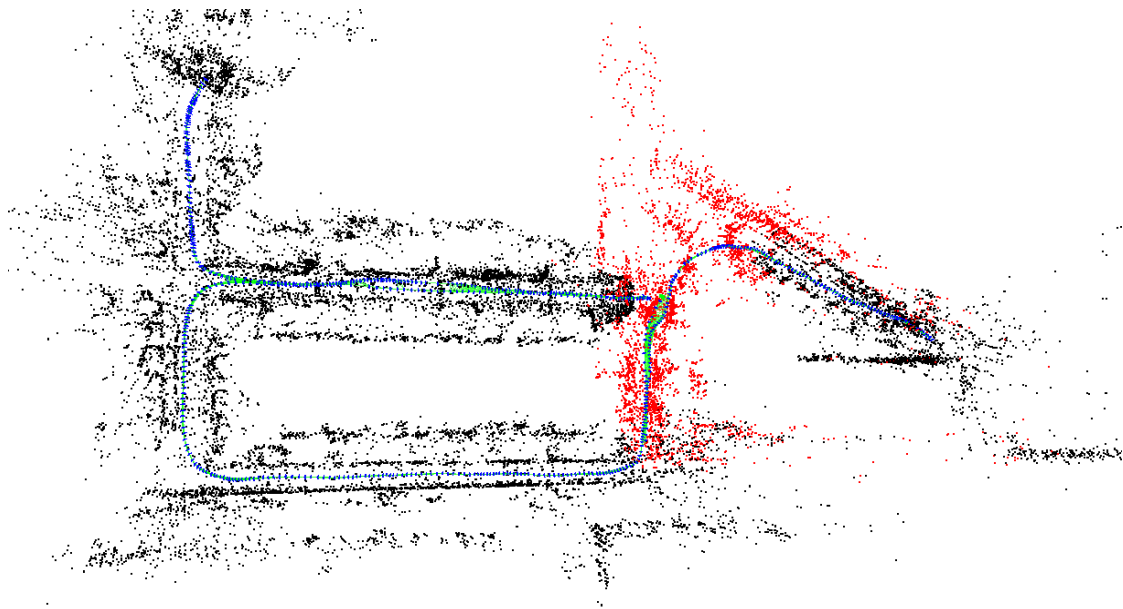

(**b**)

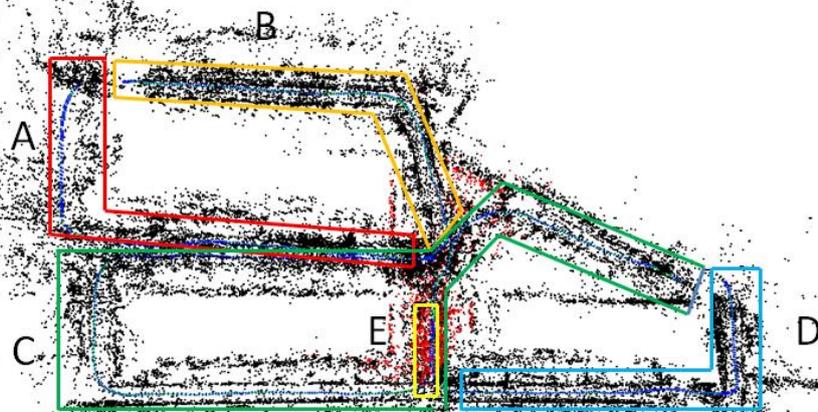

(**c**)

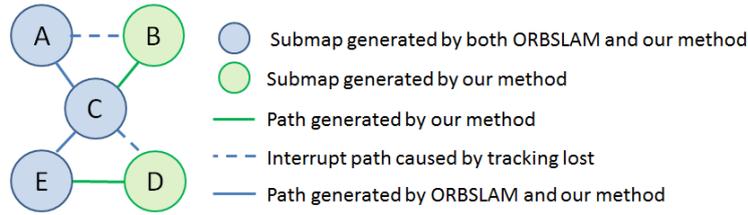

(**d**)

**Figure 14.** Outdoor street dataset experimental results. (a) The ground running trajectory. (b) The processing result of ORB-SLAM2. (c) The processing result of the proposed system (with the letters representing the submaps). (d) The undirected connected graph.

*4.3. Indoor Dataset Experiment*

4.3.1. Indoor office dataset

The third group of data is videos taken by handheld cameras, as shown in Figure 15. The video resolution is 1920 × 1080, with a total of 12,670 frames. The camera model is a GoPro Hero 6 motion camera, with a frame rate of 30 frames per second. The camera motion is more complex, some of the perspective changes dramatically, and some of the perspective is more simple. The scenes captured by this set of data are mainly of an indoor office. The scenes are relatively complex, with large white walls, which is not conducive to ORB feature extraction, and the light changes are also quite drastic.

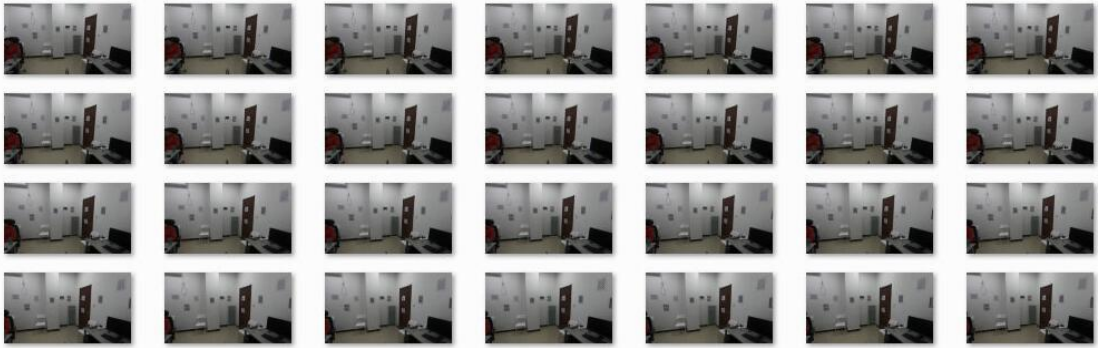

**Figure 15.** Indoor office dataset.

If the camera is shaken violently, the field of view will change greatly, leading to failure of the tracking, so there were multiple interruptions in the trajectory. In order to improve the readability, it is assumed that there are M submaps in total, in which the blue polygon box contains multiple submaps (this can be seen in Figure 16c). Figure 16a is the camera's running trajectory. Figure 16b is the processing result of ORB-SLAM2, which only reconstructs half of the room. Figure 16c is the processing result of the proposed system (with the numbers representing submaps), which reconstructs the entire room. Figure 16d is a schematic diagram of the undirected connection graph. Due to the tracking failure at the beginning of the data, ORB-SLAM2 lost most of the subsequent submaps until submap M was successfully connected to the first submap by relocation. It can be seen that the ORB-SLAM2 relocation method can easily cause a loss of map information when processing the tracking failure situation. Only when the camera captures the scene again where the previous tracking failed and maintains the tracking success state is it possible to generate a map of this part. This undoubtedly increases the workload of the mapping. The proposed method retained the first submap, and at the same time reinitialized the tracking to obtain the 2nd, 3 ... Mth connected submaps, finally obtaining the complete map.

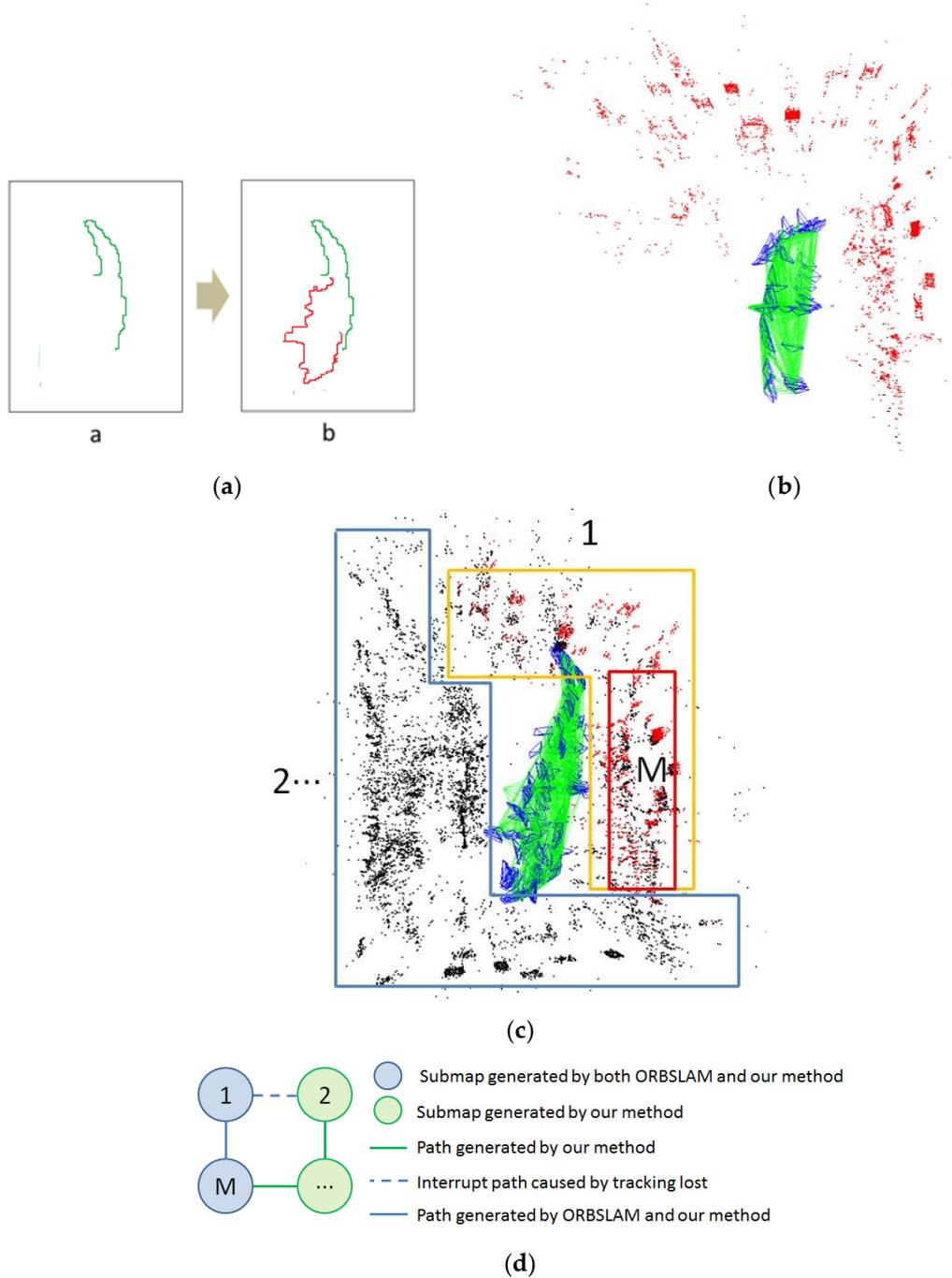

**Figure 16.** Indoor office dataset experimental results. (a) The camera's running trajectory. (b) The processing result of ORB-SLAM2. (c) The processing result of the proposed system (with the numbers representing the submaps). (d) The undirected connected graph.

4.3.2. Indoor corridor dataset

The fourth set of data has the same collection conditions as the third set, as shown in Figure 17. The video resolution is 1920 × 1080, with a total of 17,880 frames. The scenes captured by this dataset are mainly of an indoor corridor.

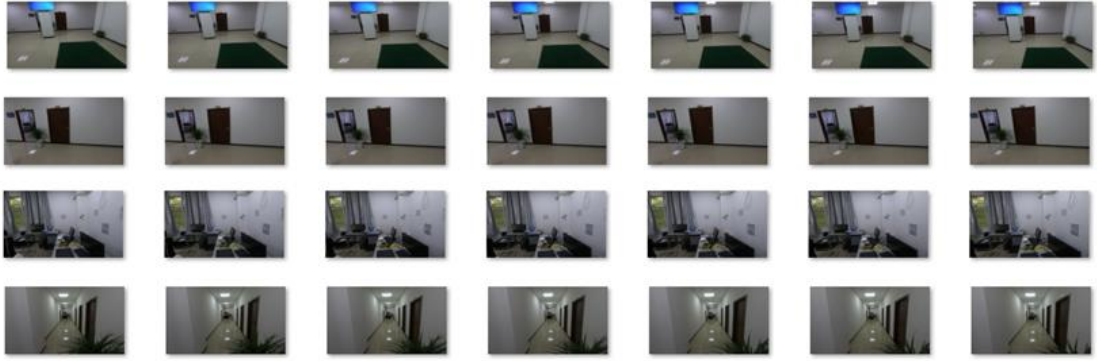

**Figure 17.** Indoor corridor dataset.

There were two interruptions in the trajectory. Figure 18a is the camera's running trajectory. Figure 18b is the processing result of ORB-SLAM2, which only reconstructs half of the corridor. Figure 18c is the processing result of the proposed system (with the letters representing the submaps), which reconstructs the entire corridor. Figure 18d is a schematic diagram of the undirected connection graph. When ORB-SLAM2 processed this set of data, it was unable to solve the position of the current frame in the world coordinate system, due to the tracking failure between submaps A and B, so submap B was lost. Submap C was successfully relocated and connected to submap A, but the final map only includes submaps A and C. After the tracking failure between the A-B submaps occurred in the proposed system, the new world coordinate system was reinitialized and the system continued tracking. At the same time, it was established whether there was a connected relationship between the submaps. Finally, the paths between submaps C and B were detected and restored, and the three submaps were connected in series with the A-C-B connected relationship.

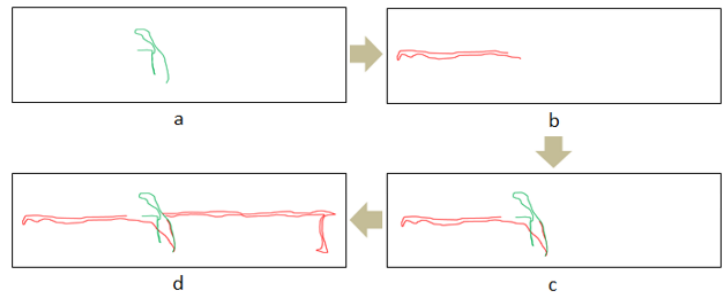

(**a**)

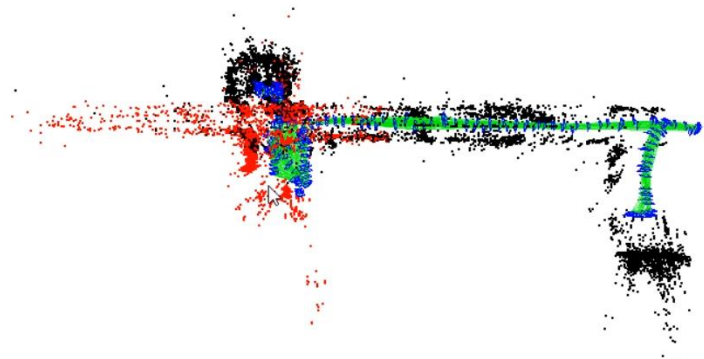

(**b**)

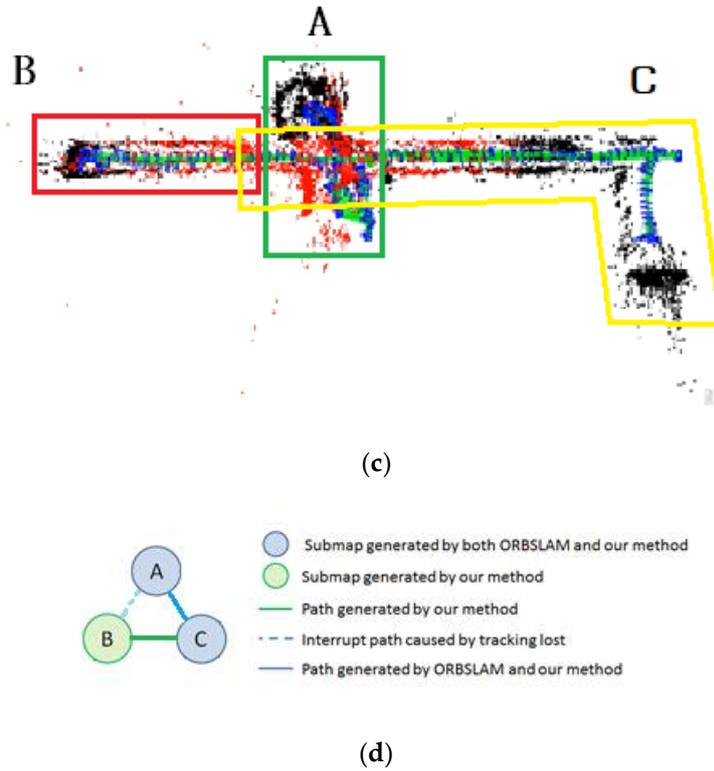

(c)

(d)

**Figure 18.** Indoor corridor dataset experimental results. (a) The camera's running trajectory. (b) The processing result of ORB-SLAM2. (c) The processing result of the proposed system (with the letters representing the submaps). (d) The undirected connected graph.

*4.4. Analysis of the Experimental Results*

The speed of the motion, sharp perspective changes, or the shooting area lacking in texture can lead to tracking failure, at which point ORB-SLAM2 will usually stop tracking, and will use the relocation method to try to restore the camera pose. However, the system cannot resume tracking until the camera returns to the location of the tracking failure. After the tracking failure, the proposed system keeps the previous submap in the stack, and reinitializes the tracking of the newly added video frame to form a new submap. When the system detects the connection between submaps, it transforms them into a whole connected map in the same coordinate system, so as to make the reconstructed map more complete. As shown in Table 2, the trajectory integrity of the proposed system is far more complete than that of ORB-SLAM2, which verifies the effectiveness of the proposed system. As shown in Table 3, from the trajectory error, the proposed system can basically keep close to the original ORB-SLAM2 level.

**Table 2.** Comparison of the trajectory integrity.

| Dataset | Trajectory integrity (number of KeyFrames retained) | |
| --- | --- | --- |
| | ORB-SLAM2 | Proposed system |
| UAV | 117 | 496 |
| Outdoor street | 296 | 465 |
| Indoor office | 30 | 68 |
| Indoor corridor | 84 | 110 |

Table 3. Comparison of the trajectory error.

| Dataset | RMSE | |
|---|---|---|
| | ORB-SLAM2(cm) | Proposed system(cm) |
| UAV | 35.421 | 36.912 |
| Outdoor street | 9.840 | 10.051 |
| Indoor office | 5.126 | 5.183 |
| Indoor corridor | 2.003 | 2.038 |

To sum up, in this study, we conducted four groups of experiments aimed at different scenarios, and the current more mainstream monocular vision SLAM framework of ORB-SLAM2 was used as an open-source comparison. The experimental results show that, in the case of tracking failure, the proposed system can rebuild a more complete scene map, confirming the effectiveness of the proposed SLAM map restoration algorithm based on submaps and an undirected connected graph.

## 5. Conclusion

In this paper, we have proposed a map information restoration algorithm based on submaps and an undirected connected graph for monocular vision SLAM. When the system fails to track multiple times and generates multiple submaps, as long as there is connectivity between the submaps, it is possible to merge the maps into a coordinate system. The proposed method is able to retain more complete map information than that retained by the existing relocation methods.

The proposed system involves starting a new submap after tracking failure occurs, and initializing the newly read video frames to track the new submap. However, because ORB-SLAM2 needs to select a pair of video frames with sufficient parallax in the initialization phase, it may need to go through more video frames before the system can be initialized successfully. To some extent, this can "waste" some video frames. Therefore, in the future, we will consider reusing the video frames during the initialization process to make the reconstructed map even more complete.

**Author Contributions:** Conceptualization, Zongqian Zhan; Funding acquisition, Zongqian Zhan; Methodology, Wenjie Jian; Software, Yihui Li; Supervision, Zongqian Zhan; Visualization, Wenjie Jian; Writing – original draft, Wenjie Jian; Writing – review & editing, Xin Wang and Yang Yue.

**Funding:** This research was funded by the National Natural Science Foundation of China (NSFC) under project number 61871295.

**Acknowledgment:** The authors would like to thank the anonymous reviewers and members of the editorial team for the comments and contributions.

**Conflicts of Interest:** The authors declare no conflict of interest.

## References

1. Davison, A.J.; Reid, I.D.; Molton, N.D.; Stasse, O. MonoSLAM: Real-time single camera SLAM. IEEE Transactions on Pattern Analysis &Machine Intelligence. 2007, 29, 1052-1067.
2. Chuang, Q.; Hui L.; Jian T.; Yu W.C.; Harri, K.; Antero, K.; Ling, L.Z.; Xin, L.L.; Liang, C.; Juha, H. An Integrated GNSS/INS/LiDAR-SLAM Positioning Method for Highly Accurate Forest Stem Mapping. Remote Sens. 2017, 9.

3. Engel, J.; Stückler, J.; Cremers, D. Large-scale direct SLAM with stereo cameras. In Proceedings of the 2015 IEEE/RSJ International Conference on Intelligent Robots and Systems (IROS), Hamburg, Germany, Sept. 28–Oct. 2, 2015; pp. 1935-1942.
4. Forster, C.; Pizzoli, M.; Scaramuzza, D. SVO: Fast semi-direct monocular visual odometry. In Proceedings of the 2014 IEEE International Conference on Robotics and Automation (ICRA), Hong Kong, PEOPLES R CHINA, MAY 31-JUN 07, 2014; pp. 15-22.
5. Mur-Artal, R.; Tardos, J.D. ORB-SLAM2: an Open-Source SLAM System for Monocular, Stereo and RGB-D Cameras. IEEE TRANSACTIONS ON ROBOTICS. 2017, 33, 1255-1262.
6. Rublee, E,; Rabaud, V.; Konolige, K.; Bradski, G. ORB: An efficient alternative to SIFT or SURF. In Proceedings of the IEEE International Conference on Computer Vision (ICCV), Barcelona, SPAIN, NOV 06-13, 2011; pp. 2564-2571.
7. Lee, T.K.; Lim, S.; Lee, S.; An, S.; Oh, S.Y. Indoor mapping using planes extracted from noisy RGB-D sensors. In Proceedings of the 25th IEEE\RSJ International Conference on Intelligent Robots and Systems (IROS), Algarve, PORTUGAL, OCT 07-12, 2012; pp. 1727-1733.
8. Taguchi, Y.; Jian, Y.D.; Ramalingam, S.; Feng, C. Point-plane SLAM for hand-held 3D sensors. In Proceedings of the IEEE International Conference on Robotics and Automation(ICRA), Karlsruhe, GERMANY, MAY 06-10, 2013; pp. 5182-5189.
9. Raposo, C.; Lourenco, M.; Barreto, J.P.; Antunes, M. Plane-based Odometry using an RGB-D Camera. In Proceedings of the 24th British Machine Vision Conference, Bristol, ENGLAND, SEP 09-13, 2013.
10. Concha, A.; Civera, J. Using Superpixels in Monocular SLAM. In Proceedings of the IEEE International Conference on Robotics and Automation(ICRA), Hong Kong, PEOPLES R CHINA, MAY 31-JUN 07, 2014; pp. 365-372.
11. Lemaire, T.; Lacroix, S. Monocular-vision based SLAM using Line Segments. In Proceedings of the IEEE International Conference on Robotics and Automation, Rome, ITALY, APR 10-14, 2007; pp. 2791-+.
12. Sola, J.; Vidal-Calleja. T.; Devy, M. Undelayed initialization of line segments in Monocular SLAM. In Proceedings of the IEEE RSJ International Conference on Intelligent Robots and Systems, St Louis, MO, OCT 10-15, 2009; pp. 1553-1558.
13. Perdices, E.; López, L.M.; Cañas, J.M. LineSLAM: Visual Real Time Localization Using Lines and UKF. In Proceedings of the ROBOT 2013: 1st Iberian Robotics Conference, November 28, 2013 - November 29, 2013; pp. 663-678.
14. Jeong, W.Y.; Lee, K.M. Visual SLAM with line and corner features. In Proceedings of the IEEE/RSJ International Conference on Intelligent Robots and Systems, Beijing, PEOPLES R CHINA, OCT 09-13, 2006; pp. 2570-2575.
15. Pumarola, A.; Vakhitov, A.; Agudo, A.; Sanfeliu, A.; Moreno-Noguer, F. PL-SLAM: Real-Time Monocular Visual SLAM with Points and Lines. In Proceedings of the 2017 IEEE International Conference on Robotics and Automation, ICRA 2017, May 29, 2017 - June 3, 2017; pp. 4503-4508.
16. Lee, Y.H.; Nam, C.; Lee, K.Y.; Li, Y.S.; Yeon, S.Y.; Doh, N.L. VPass: Algorithmic Compass using Vanishing Points in Indoor Environments. In Proceedings of the IEEE RSJ International Conference on Intelligent Robots and Systems, St Louis, MO, OCT 10-15, 2009; pp. 936-941.
17. Engel, J.; Koltun, V.; Cremers, D. Direct sparse odometry. IEEE Transactions on Pattern Analysis and Machine Intelligence. 2018, 40, 611-625.
18. Newcombe, R.A.; Lovegrove, S.; Davison, A.J. DTAM: Dense tracking and mapping in real-time. In Proceedings of the IEEE International Conference on Computer Vision (ICCV), Barcelona, SPAIN, NOV 06-13, 2011; pp. 2320-2327.
19. Forster, C.; Pizzoli, M.; Scaramuzza, D. SVO: Fast Semi-Direct Monocular Visual Odometry. In Proceedings of the IEEE International Conference on Robotics and Automation (ICRA), Hong Kong, PEOPLES R CHINA, MAY 31-JUN 07, 2014; pp. 15-22.
20. Engel, J; Schops, T.; Cremers, D. LSD-SLAM: Large-Scale Direct Monocular SLAM. In Proceedings of the 13th European Conference on Computer Vision (ECCV), Zurich, SWITZERLAND, SEP 06-12, 2014; Volume 8690, pp. 834-849.

21. Leutenegger, S.; Lynen, S.; Bosse, M.; Siegwart, R.; Furgale, P. Keyframe-Based Visual-Inertial Odometry Using Nonlinear Optimization. International Journal of Robotics Research. 2015, 34, 314-334.
22. Qin, T.; Li, P.L.; Shen, S.J. VINS-Mono: A Robust and Versatile Monocular Visual-Inertial State Estimator. IEEE TRANSACTIONS ON ROBOTICS. 2018, 34, 1004-1020.
23. Bu, S.H.;, Zhao, Y.; Wan, G.; Liu, Z.B. Map2DFusion: Real-time incremental UAV image mosaicing based on monocular SLAM. In Proceedings of the IEEE/RSJ International Conference on Intelligent Robots and Systems(IROS), Daejeon, SOUTH KOREA, OCT 09-14, 2016; pp. 4564-4571.
24. Mur-Artal, R.; Montiel, J.M.M.; Tardos, J.D. ORB-SLAM: A Versatile and Accurate Monocular SLAM System. IEEE Transactions on Robotics. 2015, 31, 1147-1163.
25. Zhan, Z.Q.; Wang, X.; Wei, M.L. Fast method of constructing image correlations to build a free network based on image multivocabulary trees. Journal of Electronic Imaging. 2015, 24.
26. Wang, X.; Zhan, Z.Q.; Heipke, C. AN EFFICIENT METHOD TO DETECT MUTUAL OVERLAP OF A LARGE SET OF UNORDERED IMAGES FOR STRUCTURE-FROM-MOTION. ISPRS Annals of the Photogrammetry. 2017, 4, 191-198.
27. Zhan, Z.Q.; Wang, C.D.; Wang, X.; Liu, Y. Optimization of incremental structure from motion combining a random k-d forest and pHash for unordered images in a complex scene. Journal of Electronic Imaging. 2018, 27.
28. Cummins, M.; Newman, P. FAB-MAP: Probabilistic Localization and Mapping in the Space of Appearance. International Journal of Robotics Research. 2008, 27, 647-665.
29. Klein, G.; Murray, D. Parallel Tracking and Mapping for Small AR Workspaces. In Proceedings of the 2007 6th IEEE and ACM International Symposium on Mixed and Augmented Reality, ISMAR, November 13, 2007 - November 16, 2007.
30. Williams, B.; Cummins, M.; Neira, J.; Newman, P.; Reid, I.; Tardos, J. An image-to-map loop closing method for monocular SLAM. In Proceedings of the 2008 IEEE/RSJ International Conference on Intelligent Robots and Systems, Nice, FRANCE, SEP 22-26, 2008; pp. 2053-+.
31. Straub, J.; Hilsenbeck, S.; Schroth, G.; Huitl, R.; Moller, A.; Steinbach, E. Fast Relocalization For Visual Odometry Using Binary Features. In Proceedings of the IEEE International Conference on Image Processing (ICIP), Melbourne, AUSTRALIA, SEP 15-18, 2013; pp. 2548-2552.
32. Moteki, A.; Yamaguchi, N.; Krasudani, A.; Yoshitake, T. Fast and accurate relocalization for keyframe-based SLAM using geometric model selection. In Proceedings of the IEEE Virtual Reality Conference (IEEE VR), Greenville, SC, MAR 19-23, 2016; pp. 235-236.
33. Ni, K.; Steedly, D.; Dellaert, F. Tectonic SAM: Exact, Out-of-Core, Submap-Based SLAM. In Proceedings of the IEEE International Conference on Robotics and Automation, Rome, ITALY, APR 10-14, 2007; pp. 1678-+.
34. Huang, S.D.; Wang, Z.; Dissanayake, G. Sparse Local Submap Joining Filter for Building Large-Scale Maps. IEEE Transactions on Robotics. 2008, 24, 1121-1130.
35. Huang, S.D.; Wang, Z.; Dissanayake, G. Exact state and covariance sub-matrix recovery for submap based sparse EIF SLAM algorithm. In Proceedings of the IEEE International Conference on Robotics and Automation, Pasadena, CA, MAY 19-23, 2008; pp. 1868-1873.
36. Tardos, J.D.; Neira, J.; Newman, P.M.; Leonard, J.J. Robust mapping and localization in indoor environments using sonar data. INTERNATIONAL JOURNAL OF ROBOTICS RESEARCH. 2002, 21, 311-330.
37. Ni, K.; Dellaert, F. Multi-level submap based SLAM using nested dissection. In Proceedings of the IEEE/RSJ International Conference on Intelligent Robots and Systems, Taipei, TAIWAN, OCT 18-22, 2010; pp. 2558-2565.
38. Zhao, L.; Huang, S.D.; Dissanayake, G. Linear SLAM: A linear solution to the feature-based and pose graph SLAM based on submap joining. In Proceedings of the IEEE/RSJ International Conference on Intelligent Robots and Systems, Tokyo, JAPAN, NOV 03-08, 2013.
39. Wang, Z.Z. Principle of photogrammetry. Wuhan university press, 2007.